\documentclass[10pt]{article} % For LaTeX2e
\usepackage[preprint]{tmlr}

\usepackage{amsmath,amsfonts,bm}

\def\eqref#1{equation~\ref{#1}}
\def\1{\bm{1}}

\DeclareMathAlphabet{\mathsfit}{\encodingdefault}{\sfdefault}{m}{sl}
\SetMathAlphabet{\mathsfit}{bold}{\encodingdefault}{\sfdefault}{bx}{n}

\usepackage{url}
\usepackage{amsfonts}
\usepackage{amsmath}
\usepackage{amssymb}
\usepackage{booktabs}
\usepackage{etoolbox}
\usepackage{graphicx}
\usepackage{latexsym}
\usepackage{multirow}
\usepackage{adjustbox}
\usepackage[frozencache,cachedir=.]{minted}
\usepackage{wrapfig}
\usepackage{caption}
\usepackage{subcaption}

\usepackage{multicol}

\usepackage{pgfplots}
\pgfplotsset{compat=newest}

\usepackage[T1]{fontenc}
\usepackage[utf8]{inputenc}

\usepackage{microtype}

\usepackage{inconsolata}

\usepackage{hyperref}
\usepackage{cleveref}

\title{ModuLoRA: Finetuning 2-Bit LLMs on Consumer GPUs by Integrating with Modular Quantizers}
\author{\name Junjie Yin \email jyin27@jhu.edu \\
        \addr Department of Computer Science \\
      \addr Johns Hopkins University
      \AND
      \name Jiahao Dong \email jd787@cornell.edu \\
        \addr Department of Computer Science \\
      \addr Cornell University and Cornell Tech
      \AND
        \name Yingheng Wang \email yw2349@cornell.edu \\
          \addr Department of Computer Science \\
      \addr Cornell University
      \AND
      \name Christopher De Sa \email cdesa@cs.cornell.edu\\
        \addr Department of Computer Science \\
      \addr Cornell University 
      \AND
      \name Volodymyr Kuleshov \email kuleshov@cornell.edu\\
        \addr Department of Computer Science \\
      \addr Cornell University and Cornell Tech }

\newcommand{\lplora}{\textsc{ModuLoRA}}
\newcommand{\llmtune}{\textsc{LLMTools}}

\newcommand{\bx}{\mathbf{x}}
\newcommand{\by}{\mathbf{y}}
\newcommand{\byi}{\mathbf{y}_i}
\newcommand{\bby}{\bar{\mathbf{y}}}
\newcommand{\bbyi}{\bar{\mathbf{y}}_i}
\newcommand{\dd}{\mathrm{d}}
\newcommand{\bA}{\mathbf{A}}
\newcommand{\bB}{\mathbf{B}}
\newcommand{\bAi}{\mathbf{A}^{(i)}}
\newcommand{\bBi}{\mathbf{B}^{(i)}} 
\newcommand{\bW}{\mathbf{W}}

\newcommand{\bWi}{\mathbf{W}^{(i)}}
\newcommand{\tbW}{\hat{\mathbf{W}}}
\newcommand{\tbWi}{\hat{\mathbf{W}}^{(i)}}
\newcommand{\bbi}{\mathbf{b}^{(i)}}
\newcommand{\bz}{\mathbf{z}}
\newcommand{\bs}{\mathbf{s}}
\newcommand{\bzi}{\mathbf{z}^{(i)}}
\newcommand{\bsi}{\mathbf{s}^{(i)}}
\newcommand{\sQ}{\mathcal{Q}}
\newcommand{\sD}{\mathcal{D}}

\newcommand{\myparagraph}[1]{\textbf{#1}\;\;}

\def\month{MM}  % Insert correct month for camera-ready version
\def\year{YYYY} % Insert correct year for camera-ready version
\def\openreview{\url{https://openreview.net/forum?id=r9p9CV52MV}} % Insert correct link to OpenReview for camera-ready version

\begin{document}

\maketitle

\begin{abstract}
We propose a memory-efficient finetuning algorithm for large language models (LLMs) that supports finetuning LLMs with 65B parameters in 2/3/4-bit precision on as little as one 24GB GPU. 
Our method, modular low-rank adaptation (\lplora), integrates any user-specified weight quantizer with finetuning via low-rank adapters (LoRAs).
Our approach relies on
a simple quantization-agnostic backward pass that adaptively materializes low-precision LLM weights from a custom black-box quantization module.
This approach enables finetuning 2-bit and 3-bit LLMs for the first time---leveraging state-of-the-art 2-bit QuIP\# quantization and 3-bit OPTQ quantization---outperforming finetuning that relies on less sophisticated 4-bit and 8-bit methods. 
In our experiments, \lplora~attains competitive performance on text classification, natural language inference, and instruction following tasks using significantly less memory than existing approaches, and we also surpass the state-of-the-art ROUGE score on a popular summarization task.
We release \lplora~together with a series of low-precision models as part of \llmtune, a user-friendly library for quantizing, running, and finetuning LLMs on consumer GPUs.
\end{abstract}

\section{Introduction}

Large language models (LLMs) excel across diverse tasks such as code generation, instruction following, and reasoning~\citep{brown2020LLMfewshot,scao2022bloom,meta2022opt}.
However, the massive size of these models---often reaching into hundreds of billions of parameters---makes them challenging to deploy on downstream tasks and motivates research into efficient finetuning algorithms \citep{li2021prefix,hulora2022}.

Here, we propose modular low-rank adaptation (\lplora), a memory-efficient finetuning algorithm for large language models (LLMs) that runs on consumer-grade hardware. 
For example, in 3-bit precision, \lplora~finetunes a LLaMA-30B model \citep{touvron2023llama} on one Nvidia RTX 3090 24GB GPU and a LLaMA-65B on one RTX A6000 48GB GPU. In 2-bit precision, \lplora~finetunes a LLaMA-30B or LLaMA-65B on one Nvidia RTX 3090 24GB GPU. 

Our approach adds high-precision low-rank adapters to the low-precision 3-bit or 4-bit weights of a frozen base LLM obtained via modern quantization algorithms \citep{hubara2021adaq,yao2021hawq3,frantar2023gptq}. 
Crucially, \lplora~does not specify its own quantization procedure---rather, it integrates with user-defined quantizers via a simple quantization-agnostic backward pass.
% It crucially relies on integrating high-precision and low-precision parameters as part of an efficient backward pass through the quantized model. 
%This pass adaptively materializes low-precision LLM weights in high precision and preserves training stability.
% Specifically, we implement an efficient 
This backward pass adaptively materializes low-precision LLM weights obtained from a black-box quantizer and integrates them with high-precision low-rank adapters. % while preserving training stability. 

We release \lplora~as part of \llmtune, a user-friendly library that enables finetuning LLMs~on consumer GPUs. 
When paired with the modern OPTQ quantizer \citep{frantar2023gptq}, \lplora~enables finetuning 3-bit LLMs for the first time, often outperforming methods based on less sophisticated 4-bit and 8-bit quantization. When paired with the state-of-the-art QuIP\# quantizer \cite{chee2023quip, albert2023quipsharp}, \lplora~ enables finetuning 2-bit LLMs for the first time, matching methods' performance on less sophisticated 4-bit and 8-bit quantization method. 
Across tasks in classification, natural language inference, and instruction following, our low-precision models achieve competitive performance 
using significantly less memory than existing approaches.
On a popular summarization benchmark, we attain a new state-of-the-art ROUGE score using a quantized LLaMA-65B model.
We open-source all our low-precision models, including the first 3-bit family of Alpaca models that feature strong instruction-following performance at multiple model sizes.
% Across our experiments, we 
% often outperform methods that rely on less sophisticated 4-bit and 8-bit quantization techniques and 
% observe only a small degradation relative to models in full precision.
% By integrating with state-of-the-art quantization algorithms, we oftentimes match and occasionally outperform simpler 8-bit finetuning algorithms, and produce the first finetuning algorithm for 3-bit LLMs.
% On a popular summarization task, our finetuned LLMs attain a new state-of-the-art ROUGE score. 
Our findings reveal that high performance can be achieved using smaller quantized LLMs than previously thought.

\textbf{Contributions.}\;
In summary, this paper makes the following contributions: (1) we propose \lplora, a memory-efficient finetuning method that operates over low-precision weights obtained via a user-specified black-box quantization module; (2) we release \llmtune, a user-friendly Python library that features an implementation of \lplora~and that enables users to easily finetune the largest LLMs on consumer GPUs; (3) we provide empirical evidence that high performance on downstream tasks can be achieved with a smaller LLM than previously thought.

\section{Background and Related Work}
\label{sec:background}

We are interested in finetuning a pre-trained LLM for downstream tasks~\citep{li2021prefix,lester2021power,houlsby2019parameter,rebuffi2017icarl}.
%Most of an LLM's learnable weights appear in $n$ linear layers
LLMs use a transformer architecture where almost all of the learnable weights---and almost all of the memory used to store these weights---appear in linear layers.\footnote{These layers include the $K$, $V$, $Q$, and $O$ projection matrices of attention blocks and the linear layers of MLP blocks.}
%we use $\bWi$ and $\bbi$ for $i \in \{ 1,2,...,n \}$ to denote their weights and biases. % of these $n$ layers.
We let the weights and biases of these $n$ linear layers be denoted $\bWi$ and $\bbi$ for $i \in \{ 1,2,...,n \}$. Given a pretrained network, our goal is to finetune it for downstream tasks using much less working memory than would be needed to store all of the $\bW$ in full precision.
% LLMs use a transformer architecture where almost all of the learnable weights---and almost all of the memory used to store these weights---appear in linear layers.
% We let the weights and biases of these $n$ linear layers be denoted $\bWi$ and $\bbi$ for $i \in \{ 1,2,...,n \}$. Given a pretrained network, our goal is to finetune it for downstream tasks. % using much less working memory than would be needed to store all of the $\bW$ in full precision.

% \subsection{Parameter-Efficient Finetuning of LLMs} Two things for this section:
% \begin{itemize}
%     \item Cite a few general fine-tuning approaches (see Sebastian Rachka blog post)
%     \item Describe LoRA in more detail.
% \end{itemize}

\subsection{Large Language Model Finetuning}

% Let's start by citing a few general fine-tuning approaches (see Sebastian Rachka blog post)
Because of the high memory requirements needed to fine-tune and store all the weights of a LLM, practitioners have developed a variety of \emph{parameter-efficient fine tuning} methods that learn in a lower dimensional space. These methods include tuning only the output layer~\citep{devlin2018bert} and tuning the prompt or prefix passed as input to an LLM~\citep{lester2021power,li2021prefix,liu2023pre,liu2023gpt}, as well as LoRA, which is the focus of this work.

%\myparagraph{LoRA.} 
\paragraph{Low-Rank Adaptation (LoRA)}
The LoRA algorithm~\citep{hulora2022} decomposes the weights $\bW$
into a sum of frozen base model weights $\bW_0 \in \mathbb{R}^{d \times d}$ and a small additive low-rank adapter $\bA \bB^\top$ consisting of the product of two rectangular matrices $\bA, \bB \in \mathbb{R}^{d \times r}$, where $r>0$ indicates the rank\footnote{For simplicity here we consider square weight matrices $\bW$; the rectangular case is a straightforward generalization.}: 
\begin{equation}
    \bW = \bW_0 + \bA \bB^\top. \label{eqn:lora}
\end{equation}
LoRA reduces the number of trained parameters by a factor of $2r/d$, lowering the storage, transmission, and task-switching overhead of inference on a system that already maintains the base model. However, LoRA must hold the base weights $\bW_0$ in memory, which requires multiple high-end GPUs
and precludes tuning large LLMs on commodity hardware.

%The computational requirements of modern machine learning models motivate a wide range of efficient machine learning algorithms [CITE].

\subsection{Low-Precision Machine Learning}

The computational requirements of modern machine learning models motivate a wide range of efficient machine learning algorithms \citep{li2021prefix,hulora2022,frantar2023gptq}.

\paragraph{Quantization}
%\myparagraph{Low-Precision Machine Learning.}
Quantization methods for neural networks reduce the number of bits required to store model weights~\citep{dong2019hawq1,dong2020hawq2,yao2022zero,park2023lutgemm}. 
% Recently OPTQ~\citep{frantar2023gptq} demonstrated the feasibility of quantizing to billion-parameter LLMs. 
A $b$-bit quantization method has the form
\begin{align}
    (\tbW_q, \bz, \bs) = \sQ(\bW) & & \tbW = \sD(\tbW_q, \bz, \bs).
\end{align}
Here, the quantization algorithm $\sQ$ takes a weight matrix $\bW \in \mathbb{R}^{d \times d}$ (or its subset) and outputs a quantized version $\tbW_q \in \{0, 1, \ldots, 2^{b-1}\}^{d \times d}$ (using $b$ bits to represent each entry of $\bW$), as well as zero and scale parameters $\bz, \bs \in \mathbb{R}^d$ (in full precision). The dequantization algorithm $\sD(\tbW_q, \bz, \bs)$ recovers an approximation $\tbW \in \mathbb{R}^{d \times d}$ by rescaling the quantized weights as $\tbW = \bs \odot \tbW_q + \bz$, where $\odot$ denotes the Hadamard product, and $\odot, +$ are extended with numpy-style broadcasting.
% \todo{VK: I would comment out what's below} To use $\tbW$ in the forward pass of a network, we can either manifest $\tbW$ in 16- or 32-bits and multiply as normal, or leverage specialized kernels to multiply using the integer values of $\tbW_q$ directly.

 Recently, \citet{frantar2023gptq} 
 proposed OPTQ, a quantization algorithm that scales to modern LLMs.
 The method iteratively runs two steps over the weight columns: (1)~quantize with nearest rounding and compute the error, (2)~update the remaining weights with a scaled error. 
 Many of our experiments finetune LLMs quantized with OPTQ.

 Following OPTQ, \citet{chee2023quip} proposed QuIP, a quantization algorithm that makes two-bit LLM compression viable for the first time. The method follows a 2-step procedure: (1)~an adaptive rounding procedure that minimizes a quadratic proxy objective,, (2)~an efficient pre- and post-processing procedure ensuring weight and Hessian incoherence through multiplication by random orthogonal matrices. Further, \citet{albert2023quipsharp} proposed QuIP\#, combining lattice codebooks with incoherence processing from QuIP to create state-of-the-art 2 bit quantized models. We show the performance of QuIP\# (with $D_4$ codebooks) quantized LLMs on the SAMSum summarization experiment. 

In concurrent work, \citet{dettmers2023qlora} proposed QLoRA, an approach for tuning quantized LLMs based on LoRA. 
While our work seeks to integrate with any user-defined quantization module (such as OPTQ), QLoRA defines its own quantization scheme, which is simpler than, say, OPTQ or QuIP. One advantage of our approach is support for 2-bit and 3-bit finetuning; QLoRA only supports 4-bit finetuning. We will also identify settings where using advanced quantizers yields performance gains over QLoRA. See Section \ref{sec:related_work} for details.
\section{Low-Precision Low-Rank Adaptation with a Modular Quantizer}
\label{sec:method}

In this section, we describe modular low-rank adaptation (\lplora), a memory-efficient finetuning algorithm for large language models (LLMs) that leverages custom quantization algorithms and runs on consumer GPU hardware.

\begin{figure}[H]
\begin{multicols}{2}
\begin{minted}[fontfamily=helvetica,fontsize=\small]{python}


class ModuLoRALinear(Module):
  """Linear ModuLoRA Layer"""
  
  def __init__(self, ...):
    self.hatWq_z_s = quantize(pretrained_W)
    (self.A, self.B) = lora_init(...)
    
  def forward(self, x):
    (hatWq, z, s) = self.hatWq_z_s
    return LPLinear.apply(x, hatWq, z, s) \
      + (x @ self.B) @ self.A.t() + self.bias

\end{minted}
\begin{minted}  [fontfamily=helvetica,fontsize=\small]{python}    


class LPLinear(Function):
  """Low-Precision Linear Map"""
  @staticmethod
  def forward(ctx, input, hatWq, z, s):
    ctx.save_for_backward(hatWq, z, s)
    hatW = dequantize(hatWq, z, s)
    output = input @ hatW.t()
    return output # hatW is deallocated
  @staticmethod
  def backward(ctx, grad_output):
    hatWq, z, s = ctx.saved_tensors
    # we recompute hatW
    hatW = dequantize(hatWq, z, s) 
    grad_input = grad_output @ hatW
    # here hatW can be deallocated
    return grad_input, None, None, None
    
\end{minted}
\end{multicols}
\caption{PyTorch pseudocode for \lplora{}.}
\label{fig:lplora_code}
\end{figure}

\subsection{Low-Rank Adaptation of Low-Precision Models}
%\paragraph{Low-Rank Adaptation in Low Precision}

%Given a pre-trained network, the first step of \lplora{} is to apply a black-box quantization method $\mathcal{Q}$ to pre-trained weights $\bWi$, yielding 
% Our approach consists in first
% The \lplora~method first 
The first step of our approach is {\em quantization}:
we apply a black-box quantization algorithm $\mathcal{Q}$ to a set of pre-trained weight matrices $\bWi$. This yields 
quantized weights, zeros, and scales $(\tbWi_q,\bzi,\bsi) = \sQ(\bWi)$. 
We use $\tbWi_q$ to denote the quantized weighs stored in low precision, while $\tbWi$ denotes the same weights materialized in high precision (both approximate the original weights $\bWi$).
Crucially, we do not specify a quantization procedure $\mathcal{Q}$ as part of \lplora---rather, we seek to support user-defined quantizers that are treated by our method is a black-box.

The core of our efforts focuses on {\em finetuning} the base quantized model.
Our method first modifies the network by replacing each linear layer---originally defined by the affine map $x \mapsto x (\bW^{(i)})^\top + \bbi$---with the reparameterized low precision \mintinline{python}{ModuLoRALinear} layer in Figure \ref{fig:lplora_code}, given by
\begin{equation}
    x \mapsto x (\tbWi)^{\top} + x \bBi (\bAi)^\top  + \bbi.
    \label{eqn:lplora_fwd}
\end{equation}
% ~defines reparameterized LoRA weight matrices $\bWi_l$ as
% \begin{equation}
%     \bWi_l = \tbWi + \bAi (\bBi)^\top,
%     \label{eqn:lplora_fwd}
% \end{equation}
Here $\bAi, \bBi \in \mathbb{R}^{d \times r}$  are learnable parameters initialized as in \citet{hulora2022}, and $\tbWi = \sD(\tbWi_q, \bzi, \bsi)$ is the fixed dequantized weight matrix. Note that this is algebraically (but not computationally) equivalent to transforming the quantized matrix as given in (\ref{eqn:lora}). 
Lastly, \lplora~fits the $\bAi$ and $\bBi$ using backprop and gradient-based learning.

% \myparagraph{Efficient Mixed-Precision Computation.}
A key challenge in this procedure is to efficiently perform computations with high-precision and low-precision tensors. Clearly, the forward pass requires multiplying by weights stored in quantized $\tbWi_q$'s. Below, we derive the backward pass for $\bAi, \bBi$ and show that it also requires multiplying by the transpose of the $\tbWi_q$'s. %above matrix.

 \subsubsection{The Structure of a Quantized Backward Pass}

 % We first introduce our approach in the context of a neural network with $n$ layers $\ffi(\bx ; \bWi_l, \bbi)$, with re-parameterized weights $\bWi_l, \bbi$ for $i = 1,2,...,n$. Given an input $\bx_0$, we use $\by_i$ to denote the output of layer $i$; the final output is $\by := \by_n$.

 % We first introduce our approach in the context of a neural network with $n$ fully-connected layers $\byi = \bWi_l \bx + \bbi$, with reparameterized weights $\bWi_l, \bbi$ and outputs $\byi$ for $i = 1,2,...,n$. The final output is $\by$.

We illustrate the technical challenges that arise in the design of a quantized backward pass in the context of a network of $n$ \mintinline{python}{ModuLoRALinear} layers. Each \mintinline{python}{ModuLoRALinear} is effectively a fully connected layer with reparameterized dense weights defined as
\begin{equation}
    \bWi_l = \tbWi + \bAi (\bBi)^\top,
    \label{eqn:lplora_fwd2}
\end{equation}
biases $\bbi$, and outputs $\byi$ for $i = 1,2,...,n$. 
 We use $\bbyi = \bWi_l \bx + \bbi$ to denote the  pre-activation output of the $i$-th step and we use $L$ to denote the loss.
 The backward pass seeks to compute gradients $\dd L / \dd \bAi$ and $\dd L / \dd \bBi$, where we overload the Leibniz notation for derivatives to also denote gradients. By the chain rule,
 \begin{equation}
     \frac{\dd L}{\dd \bAi} = \frac{\dd L}{\dd \bby_i} \cdot \frac{\dd \bby_i}{\dd \bAi}.
 \end{equation}
 Because of the additive structure of the weights $\bWi_l$ in (\ref{eqn:lplora_fwd2}), $\dd \by_i / \dd \bAi$ is straightforward to handle as it is not a function of the quantized weights $\tbWi_q$. The second term can be computed via the chain rule of calculus as
 \begin{equation}
     \frac{\dd L}{\dd \bbyi} = \frac{\dd L}{\dd \bby_{i+1}} \cdot \frac{\dd \bby_{i+1}}{\dd \byi} \cdot \frac{\dd \byi}{\dd \bbyi},
 \end{equation}
 where ${\dd \byi}/{\dd \bbyi}$ is the derivative of the activation function,
 and $\dd \bby_{i+1} / \dd \by_i = (\bWi_l)^\top = (\tbWi)^\top + \bBi (\bAi)^\top$. 
 
 The above derivations indicate that computing the gradient $\dd L / \dd \bAi$ (the argument for $\dd L / \dd \bBi$ is identical) requires performing a matrix-vector multiply $\frac{\dd L}{\dd \by_{i+1}} \cdot (\tbWi)^\top$ between a high-precision vector $\frac{\dd L}{\dd \by_{i+1}}$ with a quantized matrix $(\tbWi)^\top$. Performing this multiplication in a stable and efficient way is a challenge that we must address.
 %A key element of \lplora~is the efficient multiplication of matrices $\tbWi$ stored in low precision as $\tbWi_q$ with high-precision vectors in the forward and backward passes.

 % Show the formula for a joint backward pass. Explain which gradient is which. Explain why this is not solvable using automatic differentiation and why it requires implementing a backward pass over quantized weights.

\subsubsection{Efficient Mixed-Precision Computation of Forward and Backward Passes}

If we could precompute all dequantized weight matrices $(\tbWi)^\top$ in a high-precision format, our challenge would be solved: the matrix-vetor multiplication $\frac{\dd L}{\dd \by_{i+1}} \cdot (\tbWi)^\top$ in the backward pass would operate over two high-precision arrays, and would not introduce questions of efficiency and stability.

Unfortunately, precomputing all dequantized weight matrices $(\tbWi)^\top$ requires the same amount of GPU memory as it would take to store the original high-precision LLM.
For this computation to fit on consumer GPU hardware, we need to avoid manifesting all the $\tbWi$ in memory at once. Using (\ref{eqn:lplora_fwd}) naively, backprop would store all the $\tbWi$ from the forward pass to use them in the backward pass. 

\paragraph{Efficient Mixed Precision Computation.}
Our strategy is to \emph{recompute} the high-precision materialization $\tbWi$ of the quantized $\tbWi_q$ in the backward pass rather than save it (Figure~\ref{fig:lplora_code}).
In the \mintinline{python}{LPLinear} function, the \mintinline{python}{forward} method dequantizes $\tbWi$ and performs multiplication. Similarly, \mintinline{python}{backward} re-dequantizes $\tbWi$ and computes the gradient via dynamic programming.
% of the loss $\ell$ with respect to $x$, which is given by the chain rule from $y = x (\tbWi)^{\top}$ as
% $\nabla_x \ell = \nabla_y \ell \cdot \tbWi$.
The \mintinline{python}{hatW} 
goes out of scope and 
can be freed at the end of each method, so only one $\tbWi$ is ever stored in memory at any given time.

The amount of memory used in the forward pass of the \mintinline{python}{LPLoRA} module is small: all the intermediates are either the same size as the input $x$, or even smaller (e.g. if $x \in \mathbb{R}^{m \times d}$ then \mintinline{python}{x @ self.B} is of size $\mathbb{R}^{m \times r}$ for $r \ll d$).
The amount of additional computation involved is also small: the dequantization procedure $\tbW = \bs \odot \tbW_q + \bz$ only requires multiplying and adding a scalar to each row of $\tbW_q$.

% If we can work with the dequantized $\tbWi$ directly, finetuning is straightforward. However, in an LLM, the set of all $\tbWi$'s easily fills the memory of multiple high-end GPUs.
% A key challenge solved by \lplora~is the efficient 
% % forward and 
% gradient-based optimization of $\bAi, \bBi$ using directly the low-precision $\tbWi_q$. This enables finetuning LLMs on consumer-grade GPUs.

% \paragraph{Mixed-Precision Matrix-Vector Multiplication}

\myparagraph{Increasing Efficiency Further.}
Figure~\ref{fig:lplora_code} depicts a \emph{weight materialization} strategy in which $\tbWi$ is fully materialized at each layer in both forward and backward passes. To further reduce memory, 
% we do not need to manifest $\tbWi$ to multiply by it---our approach 
we can materialize elements of $\tbWi$ only as needed. For many quantization algorithms \citep{nagel2020up,frantar2023gptq}, we can perform \emph{row materialization}: dequantize $\tbWi$ one row at a time and immediately multiply it with an input $\bx$. % Alternatively, we can perform {\bf weight materialization}---dequantize $\tbWi$ from $\tbWi_q$, perform multiplication, and discard $\tbWi$.
\lplora{} also naturally generalizes to any direct vector-by-quantized-matrix product \emph{subroutine provided by the quantizer} $\sQ$, in which case materializing any part of $\tbWi$ may be unnecessary.
% We provide CUDA code that efficiently implements both approaches.

% For maximum efficiency, we materialize elements of $\tbWi_q$ in float16. Our base quantized LLM models are represented via weights $\tbWi_q$ (store in 3 or 4 bits), scales and zeros $\bsi, \bzi$ as well as biases $\bbi$ (all stored in 16 bits).

% Importantly, we do not need to manifest $\bW$ to multiply by it in the forward or backward passes: instead, we can directly compute
% \[
%     \bW x = \bW_0 x + \bA (\bB^\top x)
% \]
% and similarly for the backward.

% Talk about cuda kernels, Materialization in fp16 or fp32.
% Say we store the zeros and other stuff in 16bit for best results

\subsection{\llmtune: A Library for Efficient LLM Finetuning Using \lplora{}.}
We implement \lplora~as part of \llmtune, a user friendly library that enables users to 
interact with the largest LLMs on consumer hardware.
% quantize, run, and finetune the largest LLMs on consumer hardware.
The \llmtune~library enables finetuning LLMs in 2-bit, 3-bit, and 4-bit precision using the \lplora~algorithm.
It also provides an easy-to-use Python API for quantization, inference, and finetuning, as well as modular support for multiple quantizers, LLMs (including LLaMA1, LLaMA2, BLOOM, and OPT),  and optimization algorithms (including all that are compatible with the Hugging Face Trainer class).
% The \llmtune~library includes support for LLAMA1, LLAMA2, BLOOM, and OPT models (selecting a specific subset of layers to quantize and finetune), and can also support other LLMs. 
Lastly, \llmtune~supports easily loading datasets and sharing models via the HuggingFace Hub. Our code is available at: \url{https://github.com/kuleshov-group/llmtools}; our evaluation code to reproduce our results is available at: \url{https://github.com/kuleshov-group/MODULoRA-Experiment}.

A key quantization algorithm implemented in
\llmtune~is OPTQ \citep{frantar2023gptq}.
% a modern method targeted specifically at LLMs. 
% and simple nearest rounding.
In order to integrate OPTQ with LoRA-based finetuning, \llmtune~provides efficient CUDA implementations of mixed-precision matrix-vector multiplication, including row and weight materialization. 
We provide CUDA kernels for both row and weight materialization in both the forward and backward passes. 
For maximum efficiency, we materialize elements of $\tbWi_q$ in float16. The base quantized LLM models are represented via weights $\tbWi_q$ stored in $3$ or $4$ bits, with scales and zeros $\bsi, \bzi$ as well as biases $\bbi$ all stored as float16. Similarly, to integrate QuIP\# with LoRA, \llmtune~provides CUDA kernels for weight re-materialization and orthogonal matrices multiplication in the forward and backward passses. The base quantized LLM models are represented via weights $\tbWi_q$ stored in $2$ bits.

\section{Experiments}
\label{sec:experiments}

\subsection{Setup}

\myparagraph{Models.} We evaluate \lplora~and \llmtune~on the recent LLaMA~\citep{touvron2023llama} family of models, as well as open-source BLOOM \citep{scao2022bloom} and OPT models \citep{meta2022opt}. We quantize the models to 3 bits and 4 bits using OPTQ as in \citet{frantar2023gptq} with calibration 128 samples from C4 \citep{raffel2020t5}. We quantize the models to 2 bits using QuIP\# as in \citet{chee2023quip, albert2023quipsharp} with $E_8$ lattice codebooks. 

\myparagraph{Baseline.} 
We use LoRA (as implemented in the PEFT library~\citep{peft}) to finetune models quantized in 8 bits using the BitsAndBytes library~\citep{dettmers2022int8}; we also compare to full-precision results from the literature. In concurrent work, \citet{dettmers2023qlora} proposed QLoRA, a related 4-bit finetuning algorithm implemented in the BitsAndBytes library.  Accordingly, we present an experimental comparison of QLoRA with our approach, along with an in-depth discussion.

\myparagraph{Training.} 
We finetune all models on NVIDIA TITAN, 3090, and A6000 GPUs (depending on the model) with a LoRA rank of $r=8$ and alpha of $a = 32$, and report results from 3 random seeds.  We set up the training procedure following \citet{hulora2022}, with slight variation to accommodate our particular language models. For a fair comparison with the concurrent work by \citet{dettmers2023qlora}, we use the exact same hyperparameter set up. Please see Appendix \ref{sec:hyperparameter} for details on the hyperparameters used for each of our experiment. %Under these conditions, training our slowest model takes about two days. 

%%TODO: Need to update the lattice codebook. 

\subsection{Text Classification}

\myparagraph{Data \& Metrics.} We start with a simple text classification task where we seek to classify a short text snippet (up to 50 words) into its genre (e.g., fiction, telephone chat, etc.). We finetune 13B to 65B LLAMA models on 392,702 snippets from five genres and evaluate on 9,815 held out instances~\citep{N18-1101}, reporting accuracy. This yields a challenging classification task for LLMs of all sizes.

\begin{table*}[h]
\centering
% \vspace{-5mm}
% \vspace{-16pt}
% \resizebox{\textwidth}{!}{
    \begin{tabular}{l@{\hskip 0.4in}c@{\hskip 0.4in}c@{\hskip 0.4in}c}
    \toprule[1.3pt]
     LLAMA Tuning & 13B & 30B & 65B \\
    \midrule
    \llmtune~(3-bit) & 93.5 { $\pm$ 0.7} & 97.0 { $\pm$ 0.9} & 97.2 { $\pm$ 0.8} \\
    \llmtune~(4-bit) & 92.9 { $\pm$ 0.7} & 96.3 { $\pm$ 1.0} & 98.0 { $\pm$ 0.9} \\
    \midrule
    Bits\&Bytes~8-bit (LLM.int8()) & 93.0 { $\pm$ 0.7} & 93.7 { $\pm$ 1.0} & 98.6 { $\pm$ 1.0} \\
    \bottomrule
    \end{tabular}
% }
\vspace{-5pt}
\caption{Text classification accuracy (\%) for LLAMAs finetuned with LoRA \& \lplora~in 3, 4, 8 bits.}
\label{tab:text_classification}
\end{table*}

\myparagraph{Results.} 
We observe that classification accuracy consistently improves as we increase the number of parameters of the LLM. $\lplora$ combined with a 3-bit or a 4-bit LLM offers comparable performance to 8-bit finetuning in Bits\&Bytes while using significantly less memory (Table \ref{tab:text_classification}).
% The smallest 3-bit \lplora models match and sometimes exceed the performance of 8-bit models implemented in PEFT and BitsAndBytes.
% We argue that a one-shot quantizer can improve over a higher precision zero-shot quantizer like LLM.int8().

\subsection{Natural Language Inference}

\myparagraph{Data \& Metrics.} Next, we finetune LLMs on natural language inference tasks. The model is asked to predict a label from a small set (entailment, contradiction, or neutral) after being presented with a sentence pairing (a hypothesis and premise sentence pair). We finetune 7B to 65B LLaMA models on the Multi-Genre Natural Language Inference Corpus (MNLI) ~\citep{N18-1101} and evaluate on the matched test sets (in-domain examples), reporting accuracy. Baselines from GPT-3 and T5 are included, as presented in Hu et al. (2022) and ~\citet{chung2022t5}.

\myparagraph{Results.} 
% Similarly, we observe consistent increase in accuracy with increasing LLM size. 
Our 2-bit and 3-bit 65B LLaMA model matches the performance of a full-precision GPT-3+LoRA baseline. Notably, 2-bit 65B models finetuned with QuIP\# outperforms the rest of 65B models with higher precisions. We also find that \textbf{3-bit and 4-bit models from \llmtune~outperform 8-bit models from the Bits\&Bytes library for the entire model size range}.
% Notably, we find that 3-bit and 4-bit LLAMA models outperforms 8-bit LLAMA models within the 7B-30B model size range. 
2-bit, 3-bit and 4-bit \lplora~models either match or outperform their 4-bit QLoRA counterparts, often using less memory because of lower precision models. 

\begin{table*}[ht]
  \centering
    \begin{adjustbox}{max width=\textwidth}
    % \hskip 0.5cm
    \begin{tabular}{l@{\hskip 0.3in}c@{\hskip 0.2in}c@{\hskip 0.3in}c@{\hskip 0.3in}c}
    \textit{Baselines} &  &  &  & \\ 
    \toprule[1.3pt]
    Models & Finetuning Adaptation & Model Size & \# Trainable Parameters  & MNLI-m (\textit{accuracy}) \\
     % &   &   & \textit{Rouge 1/2/L}  \\
    \midrule
    GPT-3 &  Full Finetuning & 175B & 175,255.8M & 89.5 $\pm$ 0.1 \\ 
    GPT-3 & Adapter~ & 175B & 40.1M & 91.5 $\pm$ 0.1 \\ 
    GPT-3 & LoRA~ & 175B & 4.7M & 91.7 $\pm$ 0.1 \\
    \,T5 & Full Finetuning & 11B & 11,307.4M & \textbf{92.2} $\pm$ 0.1 \\
    \bottomrule   
    \end{tabular}
    \end{adjustbox}
    \vspace*{1 mm}
    \newline
    \begin{adjustbox}{max width=\textwidth}
    \begin{tabular}{lc@{\hskip 0.3in}c@{\hskip 0.35in}c@{\hskip 0.35in}c@{\hskip 0.35in}c@{\hskip 0.2in}}
    \toprule[1.3pt]
     LLaMA Finetuning & Quantizer & 7B & 13B & 30B & 65B \\
     \llmtune~(2-bit) & QuIP\#($E_8$) & 88.50 $\pm$ 0.3 & 89.72 $\pm$ 0.3  & 91.30 $\pm$ 0.3  & \textbf{91.85 $\pm$ 0.3} \\
    % \midrule
    % \llmtune~(2-bit) & QuIP\#($D_4$) & 86.59 $\pm$ 0.5 & 87.42 $\pm$ 0.5  & 89.72 $\pm$ 0.5 & 90.85  $\pm$ 0.5 \\
    \llmtune~(3-bit) & OPTQ & 88.98 $\pm$ 0.2 & 90.20 $\pm$ 0.2  & 91.09 $\pm$ 0.2 & 91.42 $\pm$ 0.1 \\
    \llmtune~(4-bit) & OPTQ & 89.31 $\pm$ 0.2 & 90.41 $\pm$ 0.2  & 91.31 $\pm$ 0.1 & 91.59 $\pm$ 0.2 \\
    \midrule
    Bits\&Bytes~(4-bit) & QLoRA & 89.28 $\pm$ 0.2 & 89.67 $\pm$ 0.2  & 91.22 $\pm$ 0.1 & 91.36 $\pm$ 0.2 \\
    Bits\&Bytes~(8-bit) & LLM.int8() & 88.95 $\pm$ 0.1 & 90.08 $\pm$ 0.1  & 91.15 $\pm$ 0.1 & 91.55 $\pm$ 0.1 \\
    \bottomrule
    \end{tabular}
    \end{adjustbox}
% }
\captionof{table}{Natural language inference on the MNLI-m dataset evaluated using classification accuracy (\%). Our LLaMA-65B-3bit model approaches state-of-the-art scores using significantly less memory.}
\label{tab:mnli-m}
\vspace{-3mm}
\end{table*}

\subsection{Abstractive Summarization}

\myparagraph{Data \& Metrics.} We finetune 7B-65B LLaMA and 7B-13B OPT models on the SAMSum dataset~\citep{gliwa2019samsum}, consisting of 14,732 (text, summary) training pairs and 819 test pairs. 
% This data features expert-written abstractive summaries of converstaions between two people.
Our methodology fully mirrors the evaluation of GPT-style models finetuned using LoRA \citep{hulora2022}.
We evaluate summarization quality using ROUGE-1/2/L; we include GPT-3 baselines from \citet{hulora2022}.

\myparagraph{Results.} 
Our 4-bit 65B LLaMA models finetuned with \lplora~outperform the GPT-3 baseline and even {\bf reach new state-of-the-art performance} on this dataset (Table~\ref{tab:summarization}).
% finetuned in full precision using both LoRA and full finetuning: even 3-bit LLAMA-13B matches this baseline (52-53\% ROUGE).
% We observe that 3-bit and 4-bit \lplora~with OPTQ 
% Additionally, 4-bit \lplora~matches or improves upon 8-bit LoRA implemented in PEFT and BitsAndBytes, again suggesting that one-shot quantization can yield improvement over zero-shot methods.
Importantly, \lplora~demonstrates performance improvements over the 4-bit QLoRA and the 8-bit BitsAndBytes methods. 
In the 7B to 65B model size range, \lplora~models (3-bit or 4-bit) outperform 8-bit LoRAs in BitsAndBytes and LLM.int8() and 4-bit LoRAs in BitsAndBytes and QLoRA. \lplora~models (2-bit) match the performance of 8-bit LoRAs in BitsAndBytes and LLM.int8() and 4-bit LoRAs in BitsAndBytes and QLoRA.
We argue that a data-driven lower precision quantization scheme can improve over a higher precision zero-shot quantizer like LLM.int8().
Switching from 4-bit to 3-bit, and then from 3-bit to 2-bit, precision within \lplora~reduces ROUGE by only about 1\%.

\begin{table*}[ht]
  \centering
    \begin{adjustbox}{max width=\textwidth}
    \begin{tabular}{l@{\hskip 0.4in}c@{\hskip 0.5in}c@{\hskip 0.5in}c}
    \textit{Baselines} &  &  & \\ 
    \toprule[1.3pt]
    Models & Finetuning Adaptation & \# Trainable Parameters  & SAMSum (\textit{Rouge 1/2/L}) \\
     % &   &   & \textit{Rouge 1/2/L}  \\
    \midrule
    GPT-3 &  Full Finetuning & 175,255.8M & 52.0 / 28.0 / 44.5 \\ 
    GPT-3 & Adapter~ & 40.1M & 53.2 / 29.0 / 45.1 \\ 
    GPT-3 & LoRA~ & 4.7M & 53.8 / 29.8 / 45.9 \\
    Pegasus & SliC~ & 2B &\textbf{ 54.4 / 29.9 / 45.9} \\
    \bottomrule
    \end{tabular}
    \end{adjustbox}
        \vspace*{1 mm}
    \newline
    \begin{adjustbox}{max width=\textwidth}
    \begin{tabular}{lc@{\hskip 0.2in}cccc}
    \toprule[1.3pt]
     LLAMA Finetuning & Quantizer & 7B & 13B & 30B & 65B \\
    \llmtune~(2-bit) & QuIP\# ($E_8$) & 51.3 / 27.3 / 43.7 & 52.3 / 29.0 / 45.0  & 53.3 / 30.2 / 46.0 & 54.0/ 30.6 / 46.2 \\
    %  \llmtune~(2-bit) & QuIP\# ($E_8$) & 51.0 / 27.6 / 43.9 & 52.1 / 29.5 / 45.2  & 53.9 / 30.7 / 46.5 & - \\
    % \midrule
    % \llmtune~(2-bit) & QuIP\# ($D_4$) & 49.2 / 26.9 / 42.7 & 50.7 / 28.6 / 44.4  & 51.6 / 30.2 / 46.4 & 52.3 / 30.5 / 46.8 \\
    \llmtune~(3-bit) & OPTQ & 51.2 / 28.2 / 44.0 & 52.4 / 29.6 / 45.1  & 53.6 / 30.8 / 46.3 & 54.1 / 30.9 / 46.5 \\
    \llmtune~(4-bit) & OPTQ & 51.7 / 28.3 / 44.4 & 53.2 / 30.2 / 46.1 & 53.9 / 31.2 / 46.9 & \textbf{54.8 / 31.3 / 47.2} \\
    \midrule
    Bits\&Bytes~(4-bit) & QLoRA & 51.6 / 28.3  / 44.5  &  51.3 / 28.1 / 44.1  &  53.0 / 30.2 / 45.7  & 53.8 / 30.5  / 45.9  \\
    Bits\&Bytes~(8-bit) & LLM.int8() & 51.9 / 28.1 / 44.5 & 51.3 / 28.2 / 43.6 & 50.8 / 28.4 / 44.1 & 53.9 / 30.4 / 46.3 \\
    \bottomrule
    \end{tabular}
    \end{adjustbox}
%}
\captionof{table}{Abstractive summarization on the SAMSum dataset evaluated using ROUGE 1/2/L. Our LLAMA-65B-4bit model obtains state-of-the-art ROUGE scores. All metrics have $\pm0.5$ confidence intervals.}
\label{tab:summarization}
\vspace{-3mm}
\end{table*}

\paragraph{Round-to-Nearest Quantization}

We also perform an ablation where we replace the OPTQ quantizer with a rount-to-nearest (RTN) approach (Table~\ref{tab:rtn}); OPTQ performs better than RTN, highlighting the importance of advanced quantizers.
\vspace{-3mm}
\paragraph{Other Model Families}

We also apply \llmtune~to the OPT \citep{meta2022opt} families of models (Table \ref{tab:summarization_opt}). Although these models perform worse than LLaMA, \lplora~matches or outperforms more memory-intensive 4-bit and 8-bit finetuning, which is consistent with our results on LLaMA.

\begin{table}[t]
    \centering
\begin{adjustbox}{max width=0.8\linewidth}
    \begin{tabular}{lccc}
    \toprule[1.3pt]
     SAMSum Performance & Quantizer & 7B & 13B \\
    \midrule
    \multirow{2}{*}{\llmtune~(3-bit)} & OPTQ  & 51.2~/~28.2~/~44.0~/~44.2 & 52.4~/~29.6~/~45.1~/~45.1 \\
     & RTN  & 50.7~/~27.2~/~43.6~/~43.6 & 51.1~/~28.7~/~44.3~/~44.5 \\
    \multirow{2}{*}{\llmtune~(4-bit)} & OPTQ  & 51.7~/~28.3~/~44.4~/~44.4 & 53.2~/~30.2~/~46.1~/~46.1 \\
     & RTN  & 51.2~/~28.5~/~44.2~/~44.2 & 52.5~/~29.9~/~45.5~/~45.5 \\
    \bottomrule
    \end{tabular}
\end{adjustbox}
\caption{OPTQ and RTN quantization with different LLaMA model sizes on the SAMSum dataset. The evaluation was done on ROUGE 1/2/L/L-Sum.}
\label{tab:rtn}
\vspace{-5mm}
\end{table}

\bigskip

\begin{table*}[h]
   \begin{minipage}[t]{\textwidth}
   \centering
    \begin{adjustbox}{max width=0.8\textwidth}
    \begin{tabular}{lc@{\hskip 0.3in}c@{\hskip 0.3in}c@{\hskip 0.05in}}
    \toprule[1.3pt]
     OPT Finetuning & Quantizer  & 13B & 30B \\
    \midrule
    \llmtune~(3-bit)& OPTQ &   48.8 / 26.7 / 41.9   & \textbf{49.9 / 27.1 / 42.5} \\
    \llmtune~(4-bit)& OPTQ &  49.3 / 26.8 / 42.0 & 49.6 / 27.1 / 42.4 \\
    \midrule
    Bits\&Bytes~(4-bit) &QLoRA &  49.2 / 27.0 / 42.1  &  49.9 / 27.0 / 42.5   \\
    Bits\&Bytes~(8-bit) &LLM.int8() &  48.8 / 26.5 / 41.7  & 49.3 / 27.1 / 42.3  \\
    \bottomrule
    \end{tabular}
    \end{adjustbox}
%}
\captionof{table}{Abstractive summarization  with OPT models on the SAMSum dataset. \lplora~in 3-bit and 4-bit precision matches ROUGE 1/2/L scores of 4-bit and 8-bit baselines. All metrics have $\pm0.5$ confidence intervals.}
\label{tab:summarization_opt}
\vspace{-5mm}
\end{minipage}
\end{table*}

\subsection{Instruction Following}

\myparagraph{Data \& Metrics.} 
We finetune 7B-65B LLaMA models on the Alpaca dataset~\citep{alpaca}, consisting 52,000 instructions, as well on the CodaAlpaca dataset~\citep{codealpaca}, consisting of 20K code generation instructions (ses \Cref{tab:code_alpaca}). We evaluate our Alpaca instruction-tuned models on the BigBenchHard (BBH) benchmark~\citep{suzgun2022challenging}, consisting of 23 challenging tasks on which LLMs do not exceed human performance. We evaluate 3-shot performance via "answer-only" prompting and use exact match accuracy as our measurement standard, testing on 6,511 samples ($\sim$ 1.5k tokens each). We include Flan and LLaMA baselines from \citet{chia2023instructeval}.%We also report ROUGE scores in the appendix.

\myparagraph{Results.} 
We find that 2-bit, 3-bit, and 4-bit performance drops only slightly relative to 8-bit models. \textbf{Crucially, 2-bit models, despite their aggressive compression, match the performance of 4-bit QLoRA in all model sizes.} 4-bit and 3-bit 65B models outperform 8-bit 30B models, despite using fewer total bits. Furthermore, 4-bit \lplora~compares well to 4-bit QLoRA, and provides consistent performance improvements, especially at smaller model sizes, where sophisticated quantization ought to provide greater benefits. This further highlights the benefits of one-shot quantization methods.
Appendix~\ref{sec:extra_experiment} also reports experiments on the CodeAlpaca dataset.

\begin{table*}[ht]
   \centering
    \begin{adjustbox}{max width=\textwidth}
    % \hskip1cm
    \begin{tabular}{llc@{\hskip 0.4in}c@{\hskip 0.3in}c@{\hskip 0.3in}c@{\hskip 0.4in}c}
    \textit{Baselines} &  &  &  &  &  \\
    \toprule[1.3pt]
    Model & Method & Quantizer & BASE (250M) & L (780M) & XL (3B) & XXL (11B) \\
    \midrule
    FLAN-T5 &  {No Finetuning} & None & 30.8 & 30.3 & 39.9  & 47.4  \\
    \bottomrule   
    \end{tabular}
\end{adjustbox}
    \vspace*{1 mm}
    \newline
\begin{adjustbox}{max width=\textwidth}
    \begin{tabular}{l@{\hskip 0.4in}l@{\hskip 0.3in}c@{\hskip 0.3in}cccc}
    \toprule[1.3pt]
    Model & Methods & Quantizer & 7B & 13B & 30B & 65B \\
    \midrule
    \multirow{6}{*}{LLaMA}
    % & {\llmtune~(2-bit)} & QuIP\# ($D_4$) & 30.3{ $\pm$ 0.7}  & 33.3{ $\pm$ 0.6}   & 37.0{ $\pm$ 0.9} & 39.3{ $\pm$ 0.9} \\
    & {\llmtune~(2-bit)} & QuIP\# ($E_8$) & 30.8{ $\pm$ 0.5}  & 33.8{ $\pm$ 0.5}   & 38.3{ $\pm$ 0.6} & 43.5{ $\pm$ 0.5} \\
     & {\llmtune~(3-bit)} & OPTQ & 31.1{ $\pm$ 0.4}  & 35.3{ $\pm$ 0.2}   & 37.2{ $\pm$ 0.6} & 43.3{ $\pm$ 0.4} \\
     & {\llmtune~(4-bit)} & OPTQ & 33.1{ $\pm$ 0.2} & 36.2{ $\pm$ 0.4} & 40.4{ $\pm$ 0.2} & 43.7{ $\pm$ 0.4} \\
    \cmidrule{2-7}
     & {Bits\&Bytes~(4-bit)} & QLoRA & 31.9{ $\pm$ 0.1} & 35.4{ $\pm$ 0.2} & 39.0{ $\pm$ 0.4} & 43.5{ $\pm$ 0.5} \\
     & {Bits\&Bytes~(8-bit)} & LLM.int8() & 33.3{ $\pm$ 0.3} & 36.8{ $\pm$ 0.2} & 39.1{ $\pm$ 0.5} & \textbf{44.7{ $\pm$ 0.4}} \\
     % & {Bits\&Bytes~16-bit} & 33.7{ $\pm$ 0.4} & 37.1{ $\pm$ 0.3} & 41.3{ $\pm$ 0.2} & 44.8{ $\pm$ 0.4} \\
     & {No Finetuning} & None & 30.9 & 37.1 & 39.3 & 42.6 \\
    \bottomrule   
    \end{tabular}
    \end{adjustbox}
% }
\captionof{table}{Instruction-tuned models evaluated on BigBench Hard (BBH). We finetune LLaMA models on the Alpaca dataset in 2 to 8 bits. We provide exact standard deviation here.}
\label{tab:bbh}
\end{table*}

\subsection{Memory Requirements} 
\begin{wrapfigure}{r}{7cm}
\centering
\vspace{-72pt} % The placement specifier can be [h!], [htb], [htbp], etc.
    \centering
\begin{tikzpicture}[scale=0.70, every node/.style={scale=1.0}] 
\begin{axis}[
	axis lines = left, % Connect the y-axis with the x-axis at the bottom
	xlabel = {Model Parameter Size (Billion)},
	ylabel = {Required Memory (Gigabyte)},
	xmin=0, xmax=70, % Adjusted for the B units
	ymode=log, % Set y-axis to logarithmic scale
	log basis y={10}, % Logarithm base 10
	log ticks with fixed point, % Display labels in fixed point notation
	ymin=1, % Set the minimum y value close to zero that can be represented on a log scale
	ymax=400, % Adjusted to include the highest y-value
	xtick={7,13,33,65},
	xticklabels={7B,13B,33B,65B},
	ytick={1,10,100,400},
	yticklabels={0GB,10GB,100GB,400GB}, % Manually set the label of the first tick to "0"
	legend style={at={(0.98,0.02)}, anchor=south east}, % Move the legend inside the plot to the bottom right
	ymajorgrids=true,
	grid style=dashed,
	every node near coord/.append style={font=\small}, % Make the font size of nodes near coords slightly bigger
]

% Data from the table for 3bit with labels
\addplot[
    color=cyan,
    mark=o,
    nodes near coords,
    point meta=explicit symbolic,
    every node near coord/.append style={yshift=0.1mm}, % Shift nodes for 3bit up a bit
    ]
    coordinates {
    (7,3) [3]
    (13,5) [5]
    (30,11) [11]
    (65,22) [22]
    };
\addlegendentry{LLMTools (2-bit)}

% Data from the table for 3bit with labels
\addplot[
    color=teal,
    mark=square,
    nodes near coords,
    point meta=explicit symbolic,
    every node near coord/.append style={yshift=1.5mm}, % Shift nodes for 3bit up a bit
    ]
    coordinates {
    (7,5) [5]
    (13,9) [9]
    (30,20) [20]
    (65,37) [37]
    };
\addlegendentry{QLoRA (4-bit)}

% \definecolor{lavender}{rgb}{0.71, 0.49, 0.86} % Defining the lavender color
% \addplot[
%     color=lavender,
%     mark=star, 
%     ]
%     coordinates {
%     (7,5)(13,9)(33,20)(65,37)
%     };
% \addlegendentry{QLoRA (4-bit)}

% Data from the table for Full precision with labels
\addplot[
    color=darkgray,
    mark=star,
    nodes near coords,
    point meta=explicit symbolic,
    ]
    coordinates {
    (7,38) [38]
    (13,74) [74]
    (30,183) [183]
    (65,360) [360]
    };
\addlegendentry{Full precision}

% Dotted line showing the difference between Full precision and 3bit
\draw [dotted, thick] (axis cs:7,5.5) -- (axis cs:7,38);
\draw [dotted, thick] (axis cs:13,9.3) -- (axis cs:13,73.9);
\draw [dotted, thick] (axis cs:30,19.4) -- (axis cs:30,183.3);
\draw [dotted, thick] (axis cs:65,35.4) -- (axis cs:65,360.4);

% Dotted line showing the difference between Full precision and 3bit
\draw [dotted] (axis cs:7,5) -- (axis cs:7,3);
\draw [dotted] (axis cs:13,9) -- (axis cs:13,5);
\draw [dotted] (axis cs:30,20) -- (axis cs:30,11);
\draw [dotted] (axis cs:65,37) -- (axis cs:65,22);

\end{axis}
\end{tikzpicture}

    \caption{Visualization of memory requirements with different methods.}
    \label{fig:memory_requirements}
% \end{figure}
\end{wrapfigure}
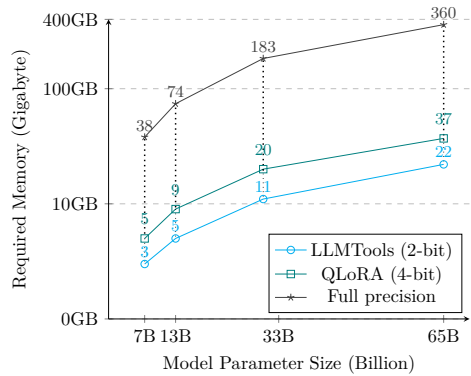

We show the memory required to perform finetuning on MNLI-M for different LLaMA model sizes in table \ref{tab:memory_requirements_table}. \lplora~ significantly minimizes the memory requirements for finetuning on these models. We plot the memory requirements in figure \ref{fig:memory_requirements} for better visualization. As the model size increases to 65B, \lplora~uses only about 6\% of the memory to run memory-efficient finetuning method LoRA. As the table and figure illustrates, with \lplora~it's possible to not only run inference but also finetune 65B model on a single 24GB GPU. To produce this table, we run our quantizer-agnostic forward/backward passes for the entire LLaMA model size range with batch size 1 and maximum sequence length 128 on MNLI-m.

\begin{table*}[ht]
  \centering
    \begin{adjustbox}{max width=0.7\textwidth}
    \begin{tabular}{l@{\hskip 0.5in}c@{\hskip 0.35in}c@{\hskip 0.35in}c@{\hskip 0.35in}c}
    \toprule[1.3pt]
     LLaMA Finetuning & 7B & 13B & 30B & 65B \\
    \midrule
    % \llmtune~(3-bit) &  5 GB  & 9 GB  & 19 GB & 35 GB \\
    \llmtune~(2-bit) & 3.2 GB & 5.4 GB  & 11.4 GB & 21.8 GB \\
    QLoRA (4-bit) & 5.2 GB & 8.6 GB  & 19.5 GB & 36.7 GB \\
    Full Precision (LoRA) & 38.4 GB & 73.9 GB  & 183.3 GB  & 360.4 GB \\
    \bottomrule
    \end{tabular}
    \end{adjustbox}
% }
\captionof{table}{Memory requirements to finetune LLaMA models on MNLI-M with batch size 1 and maximum sequence length 128. For comparison, we include the memory requirements to finetune on LoRA and QLoRA.}
\label{tab:memory_requirements_table}
% \vspace{-3mm}
\end{table*}

\section{Discussion}

\subsection{Comparison to Related Work}\label{sec:related_work}

% Want to probably add more comparison here.

\paragraph{Comparison to QLoRA}
In concurrent work, \citet{dettmers2023qlora} proposed QLoRA, a related approach for finetuning a quantized LLM. We highlight methodological and experimental differences below. 
From a methods perspective, \lplora~integrates with a user-specified black-box quantization module. In our experiments, we find that using a sophisticated data-driven quantizer like
% which is more sophisticated than zero-shot QLoRA. 
% in experiments, we find that 
OPTQ improves performance over simpler zero-shot strategies, e.g., a round-to-nearest baseline. Unlike \lplora, QLoRA defines a quantization approach similar to RTN, but also introduces a specialized packing routine, quantization of zeros and scales, and other innovations. 

From an experiments and capabilities perspective, integrating with OPTQ enables \lplora~to fintune models quantized in 2-bits and 3-bits, which QLoRA cannot do. Lastly, we identify settings where \lplora~yields LLMs with better performance than LLMs from QLoRA; this gap is likely due to the use of improved quantizers.

\paragraph{Comparison to Other Parameter-Efficient Finetuning Methods}

Recent Parameter-Efficient Finetuning (PEFT) methods have encompassed a range of techniques such as prompt tuning \citep{lester2021power, li2021prefix, qin2021learning, liu2022p}, modification of the embedding layer inputs~\citep{an2022input} or hidden states~\citep{liu2022few}, inclusion of full layers~\citep{houlsby2019parameter}, only tuning biases~\citep{zaken2021bitfit}, and others~\citep{sung2021training, karimi2021compacter}. An important shortcoming of these methods is the need to store in memory a significant amount of frozen base model parameters. 
This limits their ability to finetune the largest LLMs on consumer GPU, a limitation that we address.
% This constraint hindered their practicality for a majority of consumer-level applications: the process of computing the gradients and maintaining the optimizer states for trainable parameters still requires a considerable amount of latest-generation GPUs. Our methodology distinguishes itself by enabling efficient finetuning for most consumer-level hardwares, leveraging quantized base models. \lplora~ allows us to finetune some of the largest LLMs with just one or few GPUs. This attribute establishes the primary contrast between preceding PEFT methods and our own.

\subsection{Running LLMs on Consumer GPUs}

\paragraph{Efficient LLM Algorithms}
 The computational requirements of modern deep neural networks motivate a wide range of efficient machine learning algorithms.
 Quantization methods reduce the number of bits required to store weights~\citep{dong2019hawq1,dong2020hawq2,hubara2021adaq,li2021brecq,yao2021hawq3}, including via adaptive methods \citep{nagel2020up}.
 SmoothQuant~\citep{xiao2023smooth} rescales between activations and weights to remove outliers from the activations and make quantization overall easier.
 ZeroQuant~\citep{yao2022zero} proposes a per-layer knowledge distillation method.
 LLM.int8()~\citep{dettmers2022int8} decompose matrix multiplications into a majority of 8 bit and a minority of 16 bit operations.
 LUT-GEMM~\citep{park2023lutgemm} designs kernels to accelerate quantized matrix multiplications.
 RPTQ~\citep{yuan2023rptq} reorders activations and quantizes them in groups, reducing the impact of range differences between channels.

\paragraph{Running LLMs on Consumer GPUs}

Our methods for 3-bit and 4-bit precision enable the finetuning of a 65B LLM on a 48GB GPU, and a 30B LLM on a 24GB GPU. Additionally, our 2-bit approach allows for the finetuning of a 65B LLM on a 24GB GPU, making the finetuning of LLMs accessible on consumer hardware. Moreover, fitting an entire LLM on GPU unlocks data parallelism, which is more efficient than model parallelism. Previous 8-bit quantization methods required a 96GB GPU to fully fit a 65B model.
Finetuning GPUs on consumer hardware holds promise to accelerate model iteration and apply LLMs to a wider range of domains by a larger number of practitioners.
% Talk about memory usage before and after our method. Explain the kind of new performance you can do before and after. Talk about why 3-bits is helpful over 4-bits.

\subsection{What is a Good Base LLM for Finetuning?}

\begin{wraptable}{r}{7.5cm}
\centering
\vspace{-34pt}
{
\begin{tabular}{lccc}
\toprule[1.3pt]
Models & Quantization & BBH & PPL \\
\midrule
\multirow{2}{*}{LLAMA (13B)} & 3-bit & 35.3 & 6.63 \\
   & 4-bit & 36.2 & 5.36 \\
\midrule
\multirow{2}{*}{LLAMA (65B)} & 3-bit & 43.3 & 5.04 \\
   & 4-bit & 43.7 & 3.84 \\
\bottomrule
\end{tabular}
\vspace{-5pt}
\caption{BBH vs. PPL}
\vspace{-15pt}
}
\end{wraptable}

The traditional measure of a base LLM is perplexity. In the adjacent table, we report LLaMA perplexity (PPL) on Wiki2 as well as finetuning performance on BBH. Interestingly, the correlation is not perfect: large gaps in PPL admit small gaps in BBH.
This questions LLM evaluation when the goal is finetuning, and suggests exploring new training strategies.
% Show that worse perplexity is okay for many downstream tasks. Put figure which shows that there is correlation between preplexity and downstream tasks, but it's not perfect.

More generally, our results provide empirical evidence that high performance on downstream tasks can be achieved with a smaller quantized LLM than previously thought. While existing methods (e.g., LLM.int8()+LoRA; \citet{dettmers2022int8}) operate in 8 bits, we find that 2-bit, 3-bit, or 4-bit finetuning yields the best results for a fixed bit budget.
For example, we find that 4-bit and 3-bit 65B models outperform 8-bit and 16-bit 30B models on instruction following tasks. On the SAMSum summarization task, we find that 3-bit models are able to attain a new state-of-the-art ROUGE score, and 2-bit models match the performance of 8-bit models quantized with LLM.int8(). The high performance of these low-precision models suggests that competitive finetuning performance can be achieved on any base quantized LLM with x-bit precision, provided that the LLM exhibits reasonably good performance from the beginning.

\subsection{Limitations}
% Note: limitations doesn't count towards the page limit:
% https://2023.emnlp.org/calls/main_conference_papers/#mandatory-discussion-of-limitations
An advantage of LoRA is that it has low inference overhead, since the low-rank adaptor can be added in to the full-precision weight matrix when deploying. One limitation of \lplora~is that it does not share this advantage relative to the black-box quantized model: the low-rank adaptor cannot be trivially added to the weight matrix because the weight matrix is quantized while the adaptor is not. So, the weight matrix and adaptor cannot be fused readily, and an implementation as in Figure~\ref{fig:lplora_code} is required at inference time. A second limitation of \lplora~is that making finetuning possible on widely available commodity hardware may make finetuning too easy, presenting potential problems related to LLM safety. Another limitation of \lplora~ is that the largest models in use today (e.g. GPT-4) can have up to 1 trillion parameters, and even at the minimum of 1 bit per parameter this still would take up 125 GB, which exceeds memory on commodity GPUs: thus a straightforward application of \lplora~will be unable to make these largest-scale models finetunable on commodity hardware.

 \section{Conclusion}

Finetuning large language models typically requires substantial hardware and storage resources. Our method, \lplora, enables 2-bit finetuning of 65B models on a single 24GB consumer GPU and also supports 3-bit and 4-bit finetuning of the same models using a single 48GB GPU.  At the core of our approach is a simple, quantization-agnostic backward pass that enables integrating low-rank adapters with frozen LLM weights obtained from a user-defined quantization module. 
By integrating with modern quantizers, \lplora~achieves state-of-the-art performance compared to both parameter-efficient and full fine-tuning techniques.

\lplora's flexibility and competitive performance make finetuning more accessible and cost-effective in a resource-constrained setting.
%
% Our method, \lplora, significantly democratizes access to advanced NLP technology, enabling institutions of all sizes to compete on an even footing. 
This assists open-source model development and facilitates scientific research.
% it also opens the door to potential misuse, such as the spread of misinformation \citep{bommasani2021opportunities}. 
More broadly, we believe that \lplora~will help democratize access to large language models and make them available to a broader audience.

 \newpage

\bibliography{anthology,lora}
\bibliographystyle{tmlr}

\newpage
\appendix

\section{Additional Implementation Details}

\subsection{Configurations for BBH Evaluation}\label{app:bbh}
We evaluate the BBH dataset using LoRA adapter weights from huggingface hub with different configurations. For the Bits\&Bytes 8-bit (LLM.int8()) LoRA adapter weights, we utilized two sources: the Alpaca-7B one is obtained from the 'tloen/alpaca-lora-7b' repository, while the weights for Alpaca-13b and 30b were sourced from 'chansung/alpaca-lora-xxb'. In the case of Bits\&Bytes 4-bit (QLoRA) adapter weights, all configurations (Alpaca-7B, 13B, and 30B)—were uniformly accessed from 'timdettmers/qlora-alpaca-xxb'. Note that for the Bits\&Bytes 4-bit (QLoRA) and Bits\&Bytes 8-bit (LLM.int8()) adapter wights of the 65B model, we obtain them by finetuning the base 65B LLaMa model on Alpaca dataset using the same set of hyperparameters as ours.

\section{Additional Empirical Experiments}
\label{sec:extra_experiment}

\subsection{Additional Experiments on Code-Alpaca with LLaMA}

We conducted additional experiment on Code-Alpaca (\citep{codealpaca}). The result is shown in \Cref{tab:code_alpaca}. Consistent with our hypothesis, \lplora~performs better than or at least on par with the higher precision 8-bit models given the same number of trainable parameters and set up.

\begin{table*}[ht]
    \centering
    \begin{tabular}{p{3cm}cccc}
    \toprule
    Code Alpaca Performance & 7B & 13B & 30B & 65B \\
    \midrule
    { \llmtune~(3-bit)} & 53.6~/~36.3~/~50.7 & 57.0~/~40.0~/~53.3 & 58.1~/~40.7~/~54.3 & 60.0~/~44.1~/~58.8 \\
    { \llmtune~(4-bit)} & 54.6~/~37.2~/~51.4 & 57.4~/~40.6~/~54.3  & 59.0~/~41.4~/~57.5 & 60.2~/~43.5~/~56.8 \\
    \midrule
    { Bits\&Bytes~8-bit (LLM.int8())} & 54.0~/~36.3~/~50.9 & 57.7~/~41.3~/~54.9  & 60.6~/~43.5~/~57.5 & 61.1~/~44.1~/~58.0 \\
    \bottomrule
    \end{tabular}
% }
\caption{Instruction-tuned models evaluated using ROUGE 1/2/LSum on Code Alpaca in 3, 4, and 8 bits.}
\label{tab:code_alpaca}
\end{table*}

% \section{Additional Experiment Details and Results}
% \label{sec:extra_experiments}

% \subsection{Additional Instruction Following Experiments}

% \input{tables/bbh-app}

% \input{tables/code_alpaca}

% \input{tables/rtn}

\subsection{Finetuning \& Inference Latency}

We conducted experiment to test the finetuning and inference latency of \lplora. 

\textbf{Finetuning}. During finetuning, \lplora~significantly outperforms full-precision LoRA as show in table \ref{tab:finetuning_speed_7b}, reducing the training time by approximately 59.3\% and memory usage by 91.5\%. This efficiency in finetuning speed is primarily attributed to reduced data movement within the GPU memory.

%If we eliminate recomputation, all model weights are held in memory in full precision, and our method reduces to standard full-precision LoRA. We observe that finetuning both 4-bit and full-precision LLAMA-7B models on a single A6000 GPU results in a reduction of  training time by approximately 29.7% compared to using full-precision LoRA and in an 87.5% reduction in memory usage. ModuLoRA achieves faster finetuning speed that LoRA due to reduced data movement within the GPU memory.

\textbf{Inference}. During inference, \lplora~has a slightly lower speed compared to LoRA and QLoRA as shown in table \ref{tab:inference_speed_7b}. We attribute this to the use of CUDA kernels that are currently not as optimized as those of QLoRA. Note that 

\begin{figure}[ht]
    \begin{minipage}[t]{0.49\linewidth}
    \vskip -8.9mm
        \begin{adjustbox}{max width=\textwidth}
            \begin{tabular}{l@{\hskip 0.1in}c@{\hskip 0.1in}c@{\hskip 0.1in}c}
                \toprule[1.3pt]
                Precision & LLMTools  & QLoRA & LoRA\\
                & (2-bit)  & (4-bit) &  (Full Precision)\\
                \midrule
                Seconds/Iteration & 0.61 s/it & 0.80 s/it & 1.50 s/it \\
                \bottomrule
            \end{tabular}
        \end{adjustbox}
        \captionof{table}{Finetuning speed for LLAMA 7B on MNLI-m benchmark with batch size 1.  We report the average time to complete one step for one training data entry. To ensure fair comparison, we use a single A6000 to run on all three methods.}
        \label{tab:finetuning_speed_7b}
    \end{minipage}
    \hfill
    \begin{minipage}[t]{0.49\linewidth}
        \begin{adjustbox}{max width=\textwidth}
            \begin{tabular}{l@{\hskip 0.1in}c@{\hskip 0.1in}c@{\hskip 0.1in}c}
                \toprule[1.3pt]
                Precision & LLMTools  & QLoRA & LoRA\\
                & (2-bit)  & (4-bit) &  (Full Precision)\\
                \midrule
                Seconds/Iteration & 0.68 s/it & 0.52 s/it & 0.52 s/it \\
                \bottomrule
            \end{tabular}
        \end{adjustbox}
        \captionof{table}{Inference speed for LLAMA 7B on MNLI-m benchmark. We report the average time to complete inference for one evaluation data entry. To ensure fair comparison, we use a single A6000 to run on all three methods.}
        \label{tab:inference_speed_7b}
    \end{minipage}
\end{figure}

%1.59, 2.615, 1.95, 

% 1.96

\section{Hyperparamters Used in Experiments}
\label{sec:hyperparameter}

\subsection{LLaMA / OPT on SAMSum}

We set up the training procedure following \citet{hulora2022}, with particular accommodation to our particular language models. For a fair comparison with the concurrent work QLoRA, we use the exact same hyperparameter set up as shown in \Cref{tab:samsum_hyper} . We train using AdamW for 350 steps with a batch size of 128 samples. We report the results over 3 random seeds; the result for each run is taken from the training steps with the lowest validation loss.

\begin{table*}[h]
   \centering
\begin{adjustbox}{max width=\textwidth}
    \begin{tabular}{l@{\hskip 0.3in}l|c@{\hskip 0.5in}c@{\hskip 0.1in}}
    \toprule[1.3pt]
    Dataset & Model & LLaMA 7B / 13B / 30B / 65B & OPT 7B/ 13B / 30B  \\
    \midrule
    \multirow{10}{*}{SAMSum} & { Optimizer} & \multicolumn{2}{c}{AdamW} \\
     & { Warmup Ratio} & \multicolumn{2}{c}{0.06} \\
    \cmidrule{2-4}
     & { Batch size} & \multicolumn{2}{c}{128} \\
      & { Evaluation Batch size} & \multicolumn{2}{c}{16} \\
    & { Evaluation Steps} & \multicolumn{2}{c}{50} \\
     & { Total \# Training Steps} & \multicolumn{2}{c}{350} \\
     & { Learning Rate Schedule} & \multicolumn{2}{c}{Cosine} \\
     & { Learning Rate} & \multicolumn{2}{c}{1e-3} \\
  & { WeightDecay} & \multicolumn{2}{c}{0.0} \\
       & { LoRAConfig } & \multicolumn{2}{c}{$r_q=r_v=8$} \\
       & { LoRA $\alpha$ } & \multicolumn{2}{c}{32} \\
       & { Max Seq. Len } & \multicolumn{2}{c}{250} \\
    \bottomrule   
    \end{tabular}
    \end{adjustbox}
% }
\captionof{table}{Hyperparamters configuration for ModuLoRA, Q-LoRA on SAMSum}
\label{tab:samsum_hyper}
\end{table*}

\subsection{LLaMA on Code-Alpaca \& Text-Classification}

We again train using AdamW optimizer with a warmup ratio of 0.06. We tune learning rate, batch size, training steps for each task. We report the results over 3 random seeds. The result for each run is taken from the training steps that yield the lowest validation loss. 

\begin{table}[H]
\parbox{0.46\linewidth}{
\centering
    \begin{tabular}{l@{\hskip 0.4in}l|c}
    \toprule[1.3pt]
    Dataset & LLaMA Model & 13/30/65 B\\
    \midrule
    \multirow{9}{*}{\parbox{1.2cm}{Text-\\Classification}} & { Optimizer} & AdamW \\
     & { Warmup Ratio} & 0.06 \\
    \cmidrule{2-3}
     & { Batch size} & 256 \\
     & { Evaluation Batch size} & 32 \\
     & { Evaluation Steps} & 100 \\
     & { Total \# Training Steps} & 1000 \\
     & { Learning Rate Schedule} & Cosine \\
     & { Learning Rate} & 1e-3 \\
  & { WeightDecay} & 0.0 \\
       & { LoRAConfig } & $r_q=r_v=8$  \\
       & { LoRA $\alpha$ } & 32 \\
       & { Max Seq. Len } &  128 \\
    \bottomrule   
    \end{tabular}
\caption{Hyperparamters configuration for ModuLoRA, Q-LoRA on Text-Classification}
}
\hfill
\parbox{.47\linewidth}{
\centering
    \begin{tabular}{ll|c}
    \toprule[1.3pt]
    Dataset & LLaMA Model &  7/13/30/65 B\\
    \midrule
    \multirow{9}{*}{\parbox{1.2cm}{Code-\\Alpaca}} & { Optimizer} & AdamW \\
     & { Warmup Ratio} & 0.06 \\
    \cmidrule{2-3}
     & { Batch size} &  128 \\
     & { Evaluation Batch size} & 4 \\
     & { Evaluation Steps} & 40 \\
     & { Total \# Training Steps} & 120 \\
     & { Learning Rate Schedule} & Linear \\
     & { Learning Rate} & 1e-3\\
  & { WeightDecay} & 0.0 \\
       & { LoRAConfig } & $r_q=r_v=8$ \\
       & { LoRA $\alpha$ } & 32 \\
       & { Max Seq. Len } & 165 \\
    \bottomrule   
    \end{tabular}
\caption{Hyperparamters configuration for ModuLoRA, Q-LoRA on Alpaca-Code}
}

\end{table}

\subsection{LLaMA on MNLI-M}

Training is conducted using the AdamW optimizer, with a warmup ratio set at 0.06. We tune the learning rate, batch size, and training steps. Results are reported over three random seeds, and for each run, the performance metric is derived from the training step with the lowest validation loss. See \Cref{tab:mnli_hyper} for more details on the hyperparameters used.

\begin{table*}[ht]
   \centering
    \begin{adjustbox}{max width=0.9\textwidth}
    \begin{tabular}{l@{\hskip 0.3in}l|c@{\hskip 4.0in}c@{\hskip 4.0in}}
    \toprule[1.3pt]
    Dataset & Model & \multicolumn{2}{c}{ LLaMA 7B / 13B / 30B / 65B}\\
    \midrule
    \multirow{9}{*}{MNLI-M} & { Optimizer} & \multicolumn{2}{c}{AdamW} \\
     & { Warmup Ratio} & \multicolumn{2}{c}{0.06} \\
    \cmidrule{2-4}
     & { Batch size} & \multicolumn{2}{c}{128} \\
     & { Evaluation Batch size} & \multicolumn{2}{c}{64} \\
      & { Evaluation Steps} & \multicolumn{2}{c}{64} \\
     & { Total \# Training Epoch} & \multicolumn{2}{c}{1.0}\\
     & { Learning Rate Schedule} & \multicolumn{2}{c}{Cosine} \\
     & { Learning Rate} & \multicolumn{2}{c}{1e-3} \\
  & { WeightDecay} & \multicolumn{2}{c}{0.0} \\
       & { LoRAConfig } & \multicolumn{2}{c}{$r_q=r_v=8$} \\
       & { LoRA $\alpha$ } & \multicolumn{2}{c}{32} \\
       & { Max Seq. Len } & \multicolumn{2}{c}{128} \\
    \bottomrule   
    \end{tabular}
    \end{adjustbox}
% }
\captionof{table}{Hyperparamters configuration for ModuLoRA, Q-LoRA on MNLI-M}
\label{tab:mnli_hyper}
\end{table*}

\subsection{LLaMA on Alpaca for BBH Evaluation}

Training is conducted using the AdamW optimizer, with a warmup ratio set at 0.06. We tune the learning rate, batch size, and training steps. Results are reported over three random seeds. See \Cref{tab:bbh_hyper} for more details on the hyperparameters used.

\begin{table*}[h]
   \centering
\begin{adjustbox}{max width=\textwidth}
    \begin{tabular}{l@{\hskip 0.3in}l|c@{\hskip 0.1in}}
    \toprule[1.3pt]
    Dataset & Model & LLaMA 7B / 13B / 30B / 65B \\
    \midrule
    \multirow{9}{*}{Alpaca} & { Optimizer} & {AdamW} \\
     & { Warmup Ratio} & {0.06} \\
    \cmidrule{2-3}
     & { Batch size} & {128} \\
     & { Total \# Training Epochs} & {3} \\
     & { Learning Rate Schedule} & {Linear} \\
     & { Learning Rate} & {1e-3} \\
  & { WeightDecay} & {0.0} \\
       & { LoRAConfig } & {$r_q=r_v=8$} \\
       & { LoRA $\alpha$ } & {16} \\
       & { Max Seq. Len } & {256} \\
    \bottomrule   
    \end{tabular}
    \end{adjustbox}
% }
\captionof{table}{Hyperparamters configuration for ModuLoRA on Alpaca}
\label{tab:bbh_hyper}
\end{table*}

\end{document}